# Short Text Classification Improved by Feature Space Extension

Yanxuan Li

School of Computer Science, Beijing University of Posts and Telecommunications, Beijing, China

liyanxuanandy@163.com

**Abstract**. With the explosive development of mobile Internet, short text has been applied extensively. The difference between classifying short text and long documents is that short text is of shortness and sparsity. Thus, it is challenging to deal with short text classification owing to its less semantic information. In this paper, we propose a novel topic-based convolutional neural network (TB-CNN) based on Latent Dirichlet Allocation (LDA) model and convolutional neural network. Comparing to traditional CNN methods, TB-CNN generates topic words with LDA model to reduce the sparseness and combines the embedding vectors of topic words and input words to extend feature space of short text. The validation results on IMDB movie review dataset show the improvement and effectiveness of TB-CNN.

## 1. Introduction

As the development of mobile Internet, a large amount of meaningful short text data is generated in many ways, such as community comments, instant message applications, personal blogs, etc [1]. Therefore, how to deal with short text classification has become a problem that needs to be solved urgently. For instance, an effective classification method to distinguish malicious community comments and non-malicious ones can help create a nice user experience. However, short text has a series of features, such as shortness, sparsity, lack of semantic and contextual information [1-2]. It brings challenges for traditional methods to achieve good performance. As for reason, traditional methods of word representation, such as bag-of-word (BOW), has difficulty to express contextual information well [3]. In addition, statistical learning methods used for long documents classification need adequate word co-occurrence, such as Naive Bayes and Support Vector Machine (SVM) [1]. Moreover, it is exactly what short text lacks. Thus, these traditional methods have encountered their limitations in terms of short texts.

In related works, researches on short text mainly focus on the extension of feature space and the collection of contextual information. One way is to use external resources to enrich the feature of text. Mehran Sahami [4] and Danushka Bollegala [5] used the results returned by web search engine to measure the semantic similarity of short text. Evgeniy Gabrilovich [6] and Xiaohua Hu [7] utilized Wikipedia as an external knowledge source to compute semantic relatedness. However, with the help of external resources, these methods are quite time consuming and greatly depend on the quality of resources. Another way is to apply model-based methods, especially deep learning methods [8]. As is mentioned, word representation methods like bag-of-word does not have the ability to obtain contextual information. So Mikolov [9-11] proposed continuous bag-of-words (CBOW) model and continuous skip-gram model as a high-quality word embedding method to make up for the defects of bag-of-word model. As for classification models, Sida Wang [12] took the log-count ratio of Naive Bayes as the eigenvalue of SVM and got excellent performance in many sentiment classification datasets. Peng Wang [3] proposed a novel convolutional architecture for short text classification. Yoon Kim [13] presented a simple convolutional neural network with one-dimensional convolution layer and pre-trained word vectors, which got improvement on sentiment analysis and question classification tasks [14].

Based on the related work, we are motivated that to deal with short text classification, extending the feature space of text is an effective method. In this paper, we propose a novel topic-based convolutional neural network model (TB-CNN). The approach is an extension on convolutional neural network by extending feature space with topic words generated by LDA. We train the model on IMDB movie review dataset and compare the result with the state-of-the-art models. The experimental result shows that the proposed model obtains better performance than baseline models.

The rest of this paper is organized as follows: Section 2 describes the framework of the proposed method and how each part works. We present the evaluation result on dataset and compare the proposed model with baseline models in Section 3. At last, we summarize our work and draw conclusions in Section 4. In addition, we introduce future work in this section.

## 2. The Proposed Method

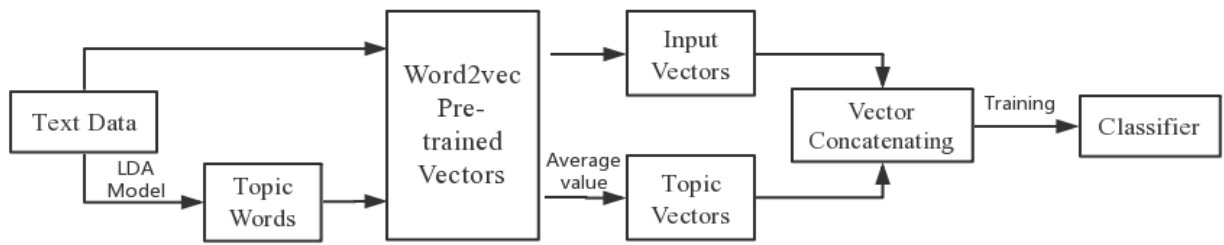

**Figure 1.** Framework of TB-CNN

The framework of the proposed approach is presented in this section, as shown in Figure 1. Firstly, we generate topic words with LDA model [15]. Next, we use pre-trained word vectors to transform input words and topic words into embedding vectors. Then, we calculate the mean of topic vectors for each topic and regard the result as the vector representation of this topic. With the input and topic vectors, we concatenate them to enrich the feature space. As for the classifier, we employ convolutional neural network and train it on IMDB dataset.

*2.1. Word2Vec*
Word2Vec is a neural network probabilistic language model proposed by Google in 2013 [9]. It embeds words with one-hot representation into a dense feature space with low dimension. So compared with traditional methods, Word2Vec model can solve the natural language processing problems with high dimensional sparse feature space [16].

The main viewpoint of Word2Vec is that words with similar contexts have similar semantics. So through training, word vector will carry contextual information. Another breakthrough achievement of Word2Vec is to provide a method to measure the similarity between words. Because the similarity between word vectors can represent their similarity in semantics, and the former can be measured by word distance such as cosine similarity.

In this paper, we use the pre-trained vector published by Google, which is trained with Google News dataset. It contains 300-dimensional vectors for 3 million words and phrases. Since the pre-trained word vector may not be suitable for specific tasks, they will be fine-tuned for our task during training.

*2.2. Latent Dirichlet Allocation*

Latent Dirichlet Allocation (LDA) is a topic generation model, also known as a three-layer Bayesian probability model [17]. It contains three-layer structure of words, topics and documents The model considers that each word in an article is obtained through a process of choosing a topic with a certain probability and then selecting a word from the topic with a certain probability. In addition, the relation between document and topic as well as the relation between topic and word obey a polynomial distribution. We employ LDA model here to identify the topic information hidden in the text. All mathematical symbols, abbreviations and their interpretations are integrated in Table 1.

In LDA model, the process of generating a document is as follows:
- Determine an article's multinomial distribution θ based on the Dirichlet distribution with parameter α.
- Sample from the polynomial distribution θ of the topic and generate the topic z of each word of the document d.
- Sample from the Dirichlet distribution with parameter β and generate the word distribution φ corresponding to the topic z.

Sample from the polynomial distribution φ of the word and finally generating the word w.

**Table 1.** Mathematical Symbols, Parameters and Their Interpretations of LDA Model.

| Name Of Symbols and Parameters | Explanation |
| --- | --- |
| D | Corpus used to train the model |
| M | Number of documents in the corpus |
| V | Number of words in the corpus |
| N | Number of words in a document |
| w | A single word, expressed as a one-dimensional vector |
| **w** | A document, **w** = ($w_1$, $w_2$, …, $w_N$), $w_n$ means the represents the nth word |
| ε | Parameter of Poisson distribution |
| α | Parameter of Dirichlet distribution |
| β | Parameter of Dirichlet distribution |
| θ | The topic distribution of a document, θ~Dirichlet(α) |
| z | Topic of a document, z~Multinomial(θ) |
| k | Number of topics |

The aim of LDA model is assigning latent topics to every document to estimate the document-topic distribution matrix and topic-word distribution matrix. The probability formula for the entire corpus is shown below.

The probability distribution formula for joint distribution of θ, w and z is formula (1). p(θ | α) is Dirichlet distribution with parameter α. p($z_n$|θ) is a multinomial distribution with parameter θ and p($w_n$|$z_n$,β) is a multinomial distribution with parameter $z_n$ and β.

$$p(\theta, z, w | \alpha, \beta) = p(\theta | \alpha) * \prod_{n=1}^{N} p(z_n | \theta) * p(w_n | z_n, \beta) \tag{1}$$

Because θ obeys a continuous distribution and z obeys a discrete distribution, for formula (1), we can integrate θ, sum z, and then multiply the generated probability formulas of all the documents in the training corpus as a joint distribution relationship.

With LDA model, we can obtain the document-topic distribution matrix, from which we can know the topic of each document. Moreover, we can find the latent keywords of each topic from the topic-word distribution matrix by ranking the probability of belonging to the topic.

*2.3. Topic-based Convolutional Neural Network (TB-CNN).*

*2.3.1. Topic Vector Concatenating.* We implement topic vector concatenating before network training. From the distribution matrixes generated by LDA model, we know the correspondence between documents and topics and the latent keywords of each topic. With pre-trained vectors of Word2Vec, we can obtain the vector representation of these keywords. For each topic, we calculate the mean of keyword vectors and regard this mean vector as the vector representation of this topic. And for each input word, we know the topic to which its document belongs. Thus, we concatenate each input word vector and its corresponding topic vector, as shown in Figure 2. The process of concatenating vectors can be summarized as shown below:
- Get the input document matrix of which the dimension is x*y. x is the number of input words of the input document, and y is the dimension of word vector.
- Find the topic words of input document with document-topic distribution matrix and transform them into word vectors.
- Calculate the mean of these word vectors as topic vector and broadcast the y-dimensional vector to an x*y topic matrix.
- Concatenate the input matrix and the topic matrix in the 1-axis dimension and obtain an x*(2y) matrix.

Vector concatenating is an important part of our approach. It represents the exploration and expansion of the short text feature space and provides a major contribution to better model performance.

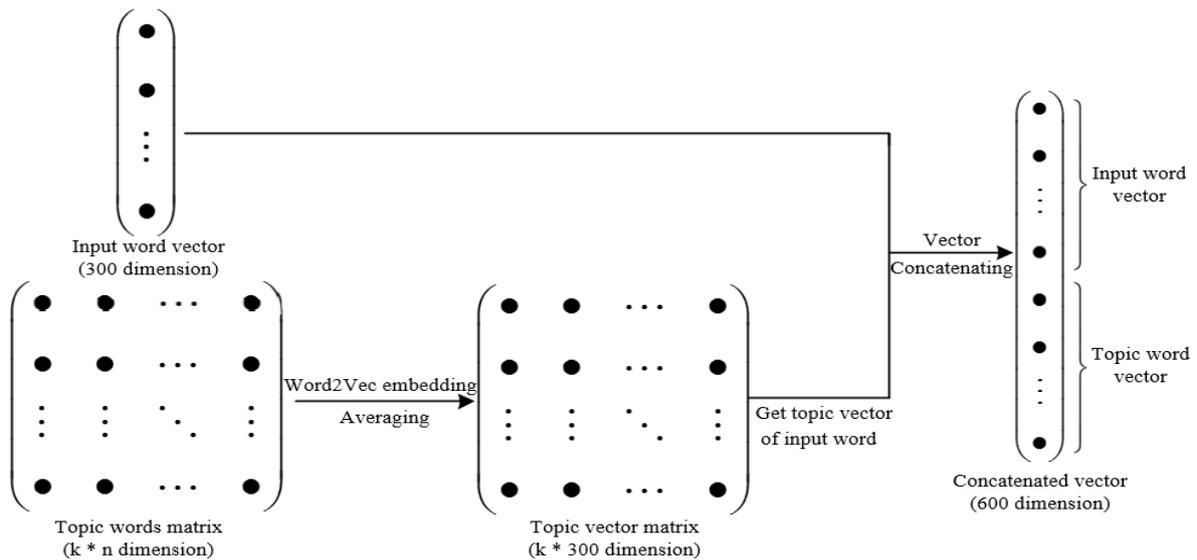

**Figure 2.** Process of topic vector concatenating

*2.3.2. Architecture of Convolutional Neural Network.* Convolutional neural networks are widely used in computer vision as part of deep learning. However, based on its characteristics, it can also be applied to the field of natural language processing. Firstly, the input layer receives a sentence of length n with necessary padding as formula (2):

$$X = \begin{bmatrix} X_1 \\ X_2 \\ \vdots \\ X_n \end{bmatrix} \quad (2)$$

We modify the embedding layer with vector concatenating mentioned in section 2.2.2, based on pre-trained vector of Word2Vec. As for convolutional layer, we perform the classification with the basic CNN architecture described by Yoon Kim [13]. We use one convolutional layer with one-dimensional convolutional kernel and multichannel. The process of a convolution operation can generate a feature map from the window of words of every filter, as shown in formula (3) and (4): (w is the parameter of convolutional layer, x represent a window of word, h is the size of window and b is a bias term)

$$c_i = f(w \cdot x_{i:i+h-1} + b) \quad (3)$$

$$c = [c_1, c_2, \cdots, c_{n-h+1}] \quad (4)$$

Then, a max-pooling layer is applied. Finally, we employ a dense layer and use softmax function as activation function to transform the score vector into a probability distribution:

$$\phi(x_i, W_z) = W_z \hat{f} \quad (5)$$

$$p(c_j | x_i, W_z) = \frac{\exp(\phi(x_i, W_z))}{\sum_{j=1}^{|C|} \exp(\phi_j(x_i, W_z))} \quad (6)$$

## 3. Experiments

*3.1. Dataset*

To validate the effectiveness of the proposed method, we conducted experiments respectively on IMDB movie review dataset. The description of dataset is shown in Table 2.

It is worth mentioning that truncation and zero-padding are applied to dataset so that the length of all input data will be 200.

**Table 2.** Description of IMDB Dataset.

| Description | Values |
|---|---|
| Size of training and testing size | 25000/25000 |
| Average number of words | 220 |
| Maximum number of words | 2361 |
| Minimum number of words | 5 |
| Classes | Positive or negative |

*3.2. LDA Model*

The main parameter that needs to be adjusted in the LDA model is the number of topics. To measure the quality of the LDA model, we use perplexity as index. The meaning of perplexity is the degree of uncertainty when a LDA model judges whether a document belongs to a topic. Therefore, a lower perplexity means a better LDA model. The perplexity of LDA model with different number of topics is shown in Table 3. As a result, we take the number of topics as 16 with lowest perplexity.

**Table 3.** Perplexity of LDA model with different number of topics.

| Number of topic | Perplexity |
| --- | --- |
| 10 | 2370.438 |
| 11 | 2374.400 |
| 12 | 2372.564 |
| 13 | 2373.258 |
| 14 | 2371.939 |
| 15 | 2369.386 |
| **16** | **2354.322** |
| 17 | 2372.139 |
| 18 | 2364.307 |
| 19 | 2360.964 |

*3.3. Pre-trained Word Vector*

In our experiments, we initialized the embedding layer with Word2Vec pre-trained word embeddings, which is publicly available. Its training corpus is Google News with every vector of 300 dimension. The size of vocabulary is 3 million, but we only load the vectors of the words found in our vocabulary and save some RAM.

*3.4. Hyper-parameters of CNN*

The design of the convolutional layer structure is the most important step in the task of convolutional neural networks. In our method, the number of convolutional layers and the dimension of the convolution kernels are already determined. Therefore, the hyper-parameter to be adjusted is the filter region size. Table 4 shows the effect of different multiple region sizes. The description of our network is shown in Table 5.

**Table 4.** Model Accuracy of CNN with Different Region Size.

| Multiple region size | Accuracy (%) |
| --- | --- |
| (2,3,4) | 91.39 |
| (3,4,5) | 91.31 |
| **(4,5,6)** | **91.40** |
| (5,6,7) | 91.19 |
| (4,4,4) | 90.86 |
| (5,5,5) | 91.01 |
| (6,6,6) | 90.80 |

**Table 5.** Description of TB-CNN.

| Description | Values |
| --- | --- |
| Pre-trained word vector | Word2Vec |
| Filter region size | (4,5,6) |
| Activation function | ReLU |
| Pooling | 1-max pooling |
| Dropout rate | 0.5 |

*3.5. Result*

Compared with the state-of-the-art methods, we introduce four famous methods as baselines. The comparison between our method and the baselines is demonstrated in Table 6 and Table 7. From the result, the proposed TB-CNN achieves the best performance with topic information added in feature space. Since we extend the feature space with our method, time for model generation and testing slightly increased.

**Table 6.** Result for Methods above. **MNB**: Multinomial Naive Bayes [12].**Bow+SVM**: Linear SVM on Bag of Words Features. **NBSVM**: SVM with Naive Bayes Features [12]. **Textcnn**: Convolutional Neural Network for Sentence Classification [13].

| Methods | Accuracy | *Precision* | Recall | *F1-score* |
|---|---|---|---|---|
| MNB | 86.59 | - | - | - |
| BoW+SVM | 87.80 | - | - | - |
| NBSVM | 91.22 | - | - | - |
| TextCNN | 88.64 | 87.32 | 90.20 | 88.58 |
| **TB-CNN** | **91.40** | **92.65** | **91.91** | **91.37** |

**Table 7.** Complexity of TextCNN and TB-CNN.

| Methods | Time for model generation and testing (s) |
|---|---|
| TextCNN | 164 |
| TB-CNN | 498 |

## 4. Conclusion

In the field of natural language processing, short text classification is an emerging and foundational task. This paper proposes a novel convolutional neural network named TB-CNN for better performance in short text classification. LDA model is used to generate topic words of each document and enrich the feature space of short text. Pre-trained word vectors are applied to generate a better representation of text with more contextual and semantic information. As a result, TB-CNN gains better performance on IMDB movie review dataset against the state-of-the-art models.

In the future, we will focus on the exploration of short text feature space. For more effective features, we will do research in how to generate topic vector with higher quality and how to add topic information into the feature space, aiming at a further improvement in short text classification.

**Acknowledgments**
My deepest gratitude goes foremost to Professor Jiali Bian, my supervisor, for her constant encouragement and guidance. She has walked me through all the stages of the writing of this paper. Without her consistent and illuminating instruction, this paper could not have reached its present form. Last my thanks would go to all those who have helped me, including my friends and my fellow classmates who gave me their help and time in listening to me and helping me work out my problems during the difficult course of the paper.

**References**
[1]   G Song, Y Ye, X Du, et al. Journal of Multimedia, 9, 635. (2014)
[2]   M Chen, X Jin, D Shen. IJCAI, 1776-1781. (2011)
[3]   P Wang, J Xu, B Xu, et al. WI-IAT, 1, 75-78. (2015)
[4]   M Sahami, TD Heilman. Proceedings of the 15th international conference on World Wide Web, 377-386. (2006)
[5]   D Bollegala, Y Matsuo, M Ishizuka. Proceedings of the 16th international conference on World

       Wide Web, 7, 757-766. (2007)
[6]    E Gabrilovich, S Markovitch. IJCAI, 7, 1606-1611. (2007)
[7]    X Hu, X Zhang, C Lu, et al. KDD, 389-396. (2009)
[8]    A Hassan, A Mahmood. ICCAR, 705-710. (2017)
[9]    T Mikolov, K Chen, G Corrado, et al. arXiv, 1301.3781. (2013)
[10]  Q Le, T Mikolov. International Conference on Machine Learning, 1188-1196. (2014)
[11]  T Mikolov, W Yih, G Zweig. NCAAL, 746-751. (2013)
[12]  S Wang, CD Manning. ACL, 90-94. (2012)
[13]  Y Kim. arXiv, 1408.5882. (2014)
[14]  DM Blei, AY Ng, MI Jordan. JMLR, 3: 993-1022. (2003)
[15]  Y Zhang, B Wallace. arXiv, 1510.03820. (2015)
[16]  F Sun, H Chen. ICIEA, 1189-1194. (2018)
[17]  Q Chen, L Yao, J Yang. ICALIP, 749-753. (2016)